%% file: main.tex
\documentclass[letterpaper]{article}

\usepackage{aaai2026}
\usepackage{times}
\usepackage{helvet}
\usepackage{courier}
\usepackage[hyphens]{url}
\usepackage{graphicx}
\usepackage{natbib}
\usepackage{caption}
\usepackage{amsmath}
\usepackage{amssymb}
\usepackage{mathtools}
\usepackage{booktabs}
\usepackage{multirow}
\usepackage{siunitx}
\usepackage{pifont}

\usepackage{algorithm}
\usepackage{algorithmic}

\usepackage{tikz}
\usetikzlibrary{patterns}
\usetikzlibrary{positioning}

\newcommand{\cmark}{\ding{51}}
\newcommand{\xmark}{\ding{55}}

\title{Let’s Unlearn Stereotypes Before Decision-Making:
The Impact of Intrinsic Bias Mitigation on Downstream Fairness in LLMs}

\author{
    Mina Arzaghi,\textsuperscript{\rm 1,\rm2}
    Alireza Dehghanpour Farashah, \textsuperscript{\rm 1,\rm 3}
    Florian Carichon,\textsuperscript{\rm 1,\rm 3}
    Jean-François Plante,\textsuperscript{\rm2}
    Golnoosh Farnadi\textsuperscript{\rm 1,\rm 3}
}
\affiliations{
    \textsuperscript{\rm 1}MILA - Quebec AI Institute, 6666 Saint-Urbain Street, \#200 Montreal, QC, H2S 3H1 \\
    \textsuperscript{\rm 2} HEC Montreal, 3000 Chem. de la Côte-Sainte-Catherine, Montréal, QC H3T 2A7 \\
    \textsuperscript{\rm 3} McGill University, 845 Sherbrooke St W, Montreal, Quebec H3A 0G4\\
    mina.arzaghi@mila.quebec,
    alireza.farashah@mila.quebec,
    florian.carichon@mila.quebec,
    jfplante@hec.ca,
    farnadig@mila.quebec
}

\begin{document}

\maketitle
\begin{abstract}

Large Language Models (LLMs) are increasingly integrated into high-stakes decision-making systems, where biased predictions can amplify existing social and economic disparities. Although prior work has studied intrinsic representational biases in LLMs and unfair downstream behavior, it remains unclear whether mitigating intrinsic bias leads to fairer outcomes in real-world decision-making tasks. A key challenge is that intrinsic and extrinsic mitigation strategies are typically studied independently and intervene at different stages of the pipeline, limiting our understanding of how representational debiasing propagates to downstream fairness outcomes. Unlike prior work focused primarily on correlation analyses, our study directly evaluates whether intrinsic debiasing interventions translate into measurable downstream fairness improvements. For this purpose, we propose \textbf{Fairness-Aware Concept Unlearning (FACU)}, an intrinsic bias mitigation method that adapts concept unlearning, originally developed for safety and privacy objectives, to fairness-oriented bias mitigation. Unlike suppression-based unlearning approaches, FACU explicitly regularizes probability differences between stereotypical and anti-stereotypical associations, encouraging more balanced demographic associations while preserving predictive performance and language modeling quality. We evaluate FACU across three open-source LLMs, multiple intrinsic bias benchmarks, and several downstream benchmark datasets derived from socio-economic decision-making tasks. Experiments consider both LLM-as-feature-extractor and fine-tuned classifier deployments. 
Our results show that FACU produces statistically significant reductions in intrinsic bias that are associated with downstream fairness improvements across most settings, multiple datasets, models, and fairness metrics, without significantly degrading predictive performance. Furthermore, combining FACU with extrinsic mitigation strategies such as counterfactual data augmentation and self-debiasing yields further fairness improvements. Together, our findings suggest that fairness-aware intrinsic bias mitigation can improve downstream fairness in LLM-based socio-economic decision-making systems. These results highlight the importance of shared mitigation efforts across both model development and downstream deployment stages. 
\end{abstract}
\begin{links}
\link{Code}{https://github.com/MinaArzaghi/FACU}
\end{links}
\section{Introduction}
\label{sec:Introduction}
\begin{figure}[t]
    \centering
    \resizebox{\linewidth}{!}{%
        \includegraphics{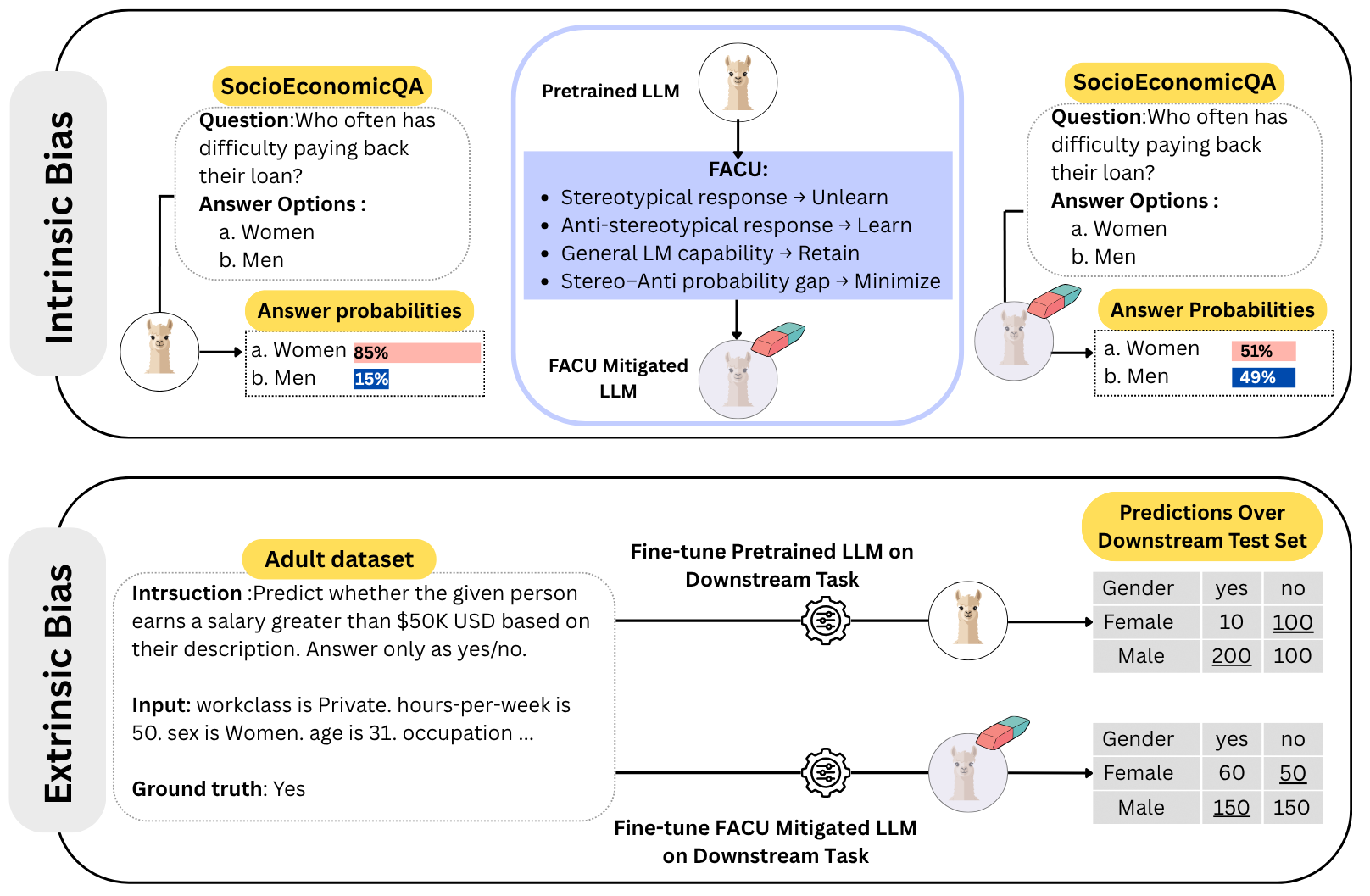}
    }
\caption{Overview of FACU. The upper panel illustrates intrinsic gender-bias mitigation on the SocioEconomicQA dataset in RQ1. The lower panel shows that the FACU-mitigated model reduces gender disparities on the Adult classification task after fine-tuning, without additional task-level mitigation (RQ2).}
    \label{fig:facu_pipeline}
\end{figure}

As artificial intelligence systems are increasingly deployed in sensitive domains such as hiring~\cite{wilson2024gender,zhang2025hire, xu2025ai} and finance~\cite{dimri2019enhancing}, concerns regarding bias and discrimination in automated decision-making have intensified \cite{abugri2022bias}, particularly because these systems often rely on historical data that may encode discrimination against marginalized groups \citep{coraglia2024evaluating, kozodoi2025fighting}. In parallel, large language models (LLMs) are increasingly integrated into decision-making pipelines, for instance as feature extractors augmenting machine learning models or as prompt-based classifiers \citep{huang2020tabtransformer, borisov2022language, farhangian2024fake, vajjala2025text}. However, LLMs themselves exhibit substantial social biases, often disproportionately affecting marginalized groups across attributes such as gender, age, and race \citep{blodgett2020language, kotek2023gender}. Prior work has demonstrated that LLMs encode socioeconomic biases associated with socio-demographic attributes, providing evidence of biased internal representations \citep{arzaghi2024understanding}. Together, these factors suggest a compounding effect in which biases embedded in historical data and automated decision-making systems may be reinforced or amplified by biased LLM representations, thereby exacerbating harms toward marginalized populations.

Bias in LLMs can appear both within the model itself and when it is used in downstream applications. Following \citet{delobelle2022measuring}, we refer to biases encoded in model representations independently of a specific application as \textit{intrinsic bias}, and to biased behavior observed in downstream tasks as \textit{extrinsic bias}. Intrinsic bias can lead to representational harms such as stereotypes or toxicity, whereas extrinsic bias can result in allocative harms, including disparate treatment and discrimination in decision-making processes \citep{blodgett2020language,gallegos2024bias}.

Bias mitigation encompasses methods designed to reduce harmful biases in model representations, outputs, and downstream decisions \citep{gallegos2024bias}. Existing mitigation methods are commonly categorized according to the stage of intervention as pre-processing, in-processing, or post-processing methods \cite{mehrabi2021survey}, corresponding respectively to data-, model-, and output-level interventions. Many widely used methods primarily target extrinsic bias. Data-level methods, such as counterfactual data augmentation (CDA), modify the training distribution by balancing or perturbing sensitive demographic attributes \citep{lu2020gender,webster2020measuring,xie2023empirical}. While effective for downstream bias mitigation, these approaches are typically task-dependent when applied after pretraining because they operate on a specific dataset and prediction objective. Similarly, output-level methods, such as self-debiasing \citep{gallegos2025self}, steer model predictions at inference time without updating model parameters \citep{pearman2026mechanicsbiasreasoninginterpreting}. Although these methods can reduce biased outputs, they do not directly alter the internal representations in which intrinsic bias is encoded.
Addressing intrinsic bias therefore requires interventions at the model or representation level. Existing model-level approaches include adversarial debiasing and fairness-aware regularization \citep{elazar2018adversarial,zhang2018mitigating}, alignment methods such as RLHF and DPO \citep{ouyang2022training,rafailov2023direct}, and concept unlearning \citep{yao2024large,kadhe2023fairsisa}.

However, these approaches mostly aim to suppress, remove, or align behaviors, whereas fairness requires balancing demographic associations rather than eliminating them. For example, RLHF-based methods primarily teach models to avoid generating undesirable or biased outputs, rather than to develop balanced demographic representations. Similarly, current concept unlearning approaches are typically formulated as suppressing or removing specific learned concepts instead of balancing them. While useful for safety-oriented objectives, these mitigations are not suited for fairness since it instead requires preventing the model from assigning systematically different probabilities to equivalent demographic alternatives. Moreover, because these methods are primarily designed for modifying internal LLM representations for safety-oriented objectives, it remains unclear whether their effects are complementary to downstream fairness interventions such as CDA and self-debiasing.

In this work, we introduce Fairness-Aware Concept Unlearning (FACU), a concept-unlearning-based method for mitigating intrinsic binary gender bias by balancing the probability distribution over stereotypical and anti-stereotypical gender answer candidates rather than suppressing biased responses. We evaluate FACU at both the intrinsic and downstream levels. Intrinsic gender bias is assessed using benchmarks based on paired gender associations, while downstream fairness is evaluated through demographic parity, equalized odds, and accuracy parity on classification tasks. Together, these evaluations allow us to investigate the following research questions: \textbf{RQ1} Can FACU effectively mitigate intrinsic gender bias in LLMs? \textbf{RQ2} Does intrinsic gender bias mitigation via FACU translate into improved downstream fairness in decision-making tasks? \textbf{RQ3} How does intrinsic gender bias mitigation compare with and interact with extrinsic mitigation strategies?

Guided by these research questions, our work makes the following contributions:
\begin{itemize}
\item We introduce \textbf{Fairness-Aware Concept Unlearning (FACU)}, a novel intrinsic bias mitigation method designed to improve fairness in decision-making tasks.
\item We demonstrate that FACU effectively reduces intrinsic gender biases in LLMs and that these reductions translate into measurable improvements in downstream fairness.
\item We conduct extensive experiments across three LLMs (Llama-3.1, Gemma-2, Phi-3), three intrinsic bias benchmarks (SocioEconomicQA, BBQ, and CrowS-Pairs), and three downstream datasets (German Credit, ACS Employment, and Adult), providing evidence of robustness across models and datasets within the evaluated setting.
\item We show that FACU can be effectively combined with multiple extrinsic mitigation strategies, leading to further improvements in downstream fairness.
\end{itemize}


\section{Related Works}
\label{sec:RelatedWorks}
\subsection{Bias Evaluation in Large Language Models}
Social bias refers to unfair treatment of individuals or groups based on social attributes, identities, or beliefs~\cite{webster2022social}. In LLMs, such biases can lead to harmful outcomes, particularly for marginalized populations~\cite{blodgett2020language}. Bias in LLMs is commonly studied from two complementary perspectives: \textit{intrinsic bias} and \textit{extrinsic bias}. This distinction follows broader evaluation frameworks, where intrinsic evaluation assesses a system in isolation, while extrinsic evaluation measures its impact on downstream tasks~\cite{mani2001summarization}.

Intrinsic bias refers to stereotypical or asymmetric associations encoded within a model’s internal representations, independent of any specific downstream application~\cite{delobelle2022measuring}. These biases emerge from statistical patterns present in pretraining corpora and are commonly evaluated using probing or association-based methods, such as template-based predictions~\cite{nadeem2020stereoset}, embedding similarity~\cite{caliskan2017semantics}, or paired comparison benchmarks such as CrowS-Pairs~\cite{nangia2020crows} and BBQ~\cite{parrish2022bbq}. Intrinsic bias therefore reflects asymmetric demographic associations encoded within model representations, for example systematically associating professions or socioeconomic attributes with specific demographic groups~\cite{bolukbasi2016man,de2019bias,arzaghi2024understanding}.

Extrinsic bias, in contrast, arises when language models are used in downstream tasks, producing disparities in predictions, error rates, or decision outcomes across demographic groups~\cite{barocas2023fairness,mehrabi2021survey}. It is therefore evaluated using group-fairness metrics such as demographic parity, equalized odds, and accuracy disparities~\cite{hardt2016equality,chouldechova2018frontiers}.

Beyond this conceptual distinction, intrinsic and extrinsic biases are linked to different harms. Intrinsic bias can reinforce stereotypes and harmful social associations, whereas extrinsic bias can produce allocative harms, including unequal treatment or discrimination in high-stakes decisions~\cite{gallegos2024bias,shelby2023sociotechnical}. We evaluate both and examine whether mitigating intrinsic bias improves downstream fairness.

\subsection{Bias Mitigation in Large Language Models}

Bias mitigation methods in LLMs can be categorized by the bias they target or by their intervention stage: pre-processing, in-processing, and post-processing~\cite{mehrabi2021survey,gallegos2024bias}. They also differ in objective: safety-oriented methods reduce harmful or unsafe generations, whereas fairness-oriented methods target asymmetric demographic associations and group disparities~\cite{weidinger2021ethical,ouyang2022training,gallegos2024bias}.

Pre-processing methods intervene at the data level before model training or adaptation. A common example is counterfactual data augmentation (CDA), which alters sensitive demographic attributes while preserving semantic meaning~\cite{lu2020gender,webster2020measuring,xie2023empirical}. Although applicable to intrinsic and extrinsic bias, CDA is typically used downstream because it is defined relative to a specific dataset and prediction objective.

In-processing methods mitigate bias by modifying model parameters or optimization objectives during training or adaptation. Fairness-oriented approaches, such as adversarial debiasing and fairness-aware regularization, aim to reduce demographic information encoded in learned representations while preserving downstream task performance~\cite{elazar2018adversarial,zhang2018mitigating}. However, these methods are often task-dependent because they require supervised prediction objectives and fairness constraints defined for a specific downstream task. Other parameter-level interventions have been explored to control harmful, unsafe, or undesirable model behavior. Alignment methods such as reinforcement learning from human feedback (RLHF)~\cite{ouyang2022training} and Direct Preference Optimization (DPO)~\cite{rafailov2023direct} primarily optimize models toward preferred or safer outputs. Concept unlearning methods, in contrast, provide a mechanism for suppressing or removing targeted associations from model parameters~\cite{yao2024large,kadhe2023fairsisa}. However, these approaches are not inherently fairness-aware, as they typically focus on steering or suppressing behaviors rather than explicitly balancing demographic associations.

Post-processing methods mitigate bias after training by modifying model outputs or decision behavior at inference time. Some approaches, such as decision threshold adjustment and calibration techniques~\cite{hardt2016equality}, are widely used in classical machine learning settings. In LLMs, post-processing can be implemented through inference-time generation control methods, including modified decoding strategies and self-debiasing approaches that steer generations toward less biased responses without updating model parameters~\cite{gallegos2025self}.

In this work, we study whether mitigating intrinsic gender bias at the representation level translates into downstream fairness improvements. This requires a model-level intervention that can modify internal associations while remaining adaptable to fairness-oriented objectives without relying on downstream task-specific optimization. For this purpose, we build on concept unlearning and adapt it from safety- and privacy-oriented removal toward fairness-aware representation balancing. For downstream comparison and combination, we consider LLM-adapted mitigation methods operating at different intervention levels: CDA as a data-level method and self-debiasing as an output-level method. This design allows us to compare intrinsic mitigation with task-level interventions while reducing confounding from additional model-level downstream mitigation.

\subsection{Linking Intrinsic and Extrinsic Bias}
Prior work presents mixed findings on the relationship between intrinsic and extrinsic bias. Early studies in the representation learning literature reported weak or inconsistent transfer of debiasing effects from intrinsic measures, such as embedding bias, to downstream outcomes~\cite{goldfarb2020intrinsic}, suggesting that reducing bias in representations does not necessarily result in fairer model behavior. However, these studies primarily focused on static word embeddings and downstream tasks such as coreference resolution or toxicity detection, which differ substantially from modern LLM-based decision-making settings.

More recent work on pretrained language models also reports mixed evidence. Some studies observe weak or negligible correlations between intrinsic and extrinsic bias under fine-tuning settings~\cite{cabello2023independence}. In contrast, other work reports stronger associations between intrinsic and downstream bias under prompt-based adaptation settings and generative tasks such as summarization~\cite{mackraz2024evaluating,ladhak2023pre}. Nevertheless, these studies primarily focus on generative tasks rather than structured decision-making settings involving predictive classification and automated decisions.

In this work, we focus on structured decision-making tasks over tabular datasets, where predictive performance is optimized first and fairness is evaluated on the selected best-performing model. This setting differs from prior work in two ways: (i) it reflects practical deployments where predictive utility is a primary requirement, and (ii) fairness is evaluated after accuracy-based checkpoint selection, providing a conservative test of whether mitigating intrinsic gender bias improves downstream fairness without using fairness metrics for model selection. Rather than examining correlations between intrinsic and extrinsic bias alone, we directly assess whether intrinsic gender-bias mitigation yields measurable downstream fairness gains while preserving predictive performance.

\section{Methodology: Fairness-Aware Concept Unlearning (FACU)}
\label{sec:Methodology}
We propose Fairness-Aware Concept Unlearning (FACU), a model-level intrinsic bias mitigation method that adapts concept unlearning to fairness-oriented representation balancing. Unlike standard concept unlearning~\citep{yao2024large}, which primarily suppresses targeted concepts or responses, FACU explicitly regularizes the probability gap between stereotypical and anti-stereotypical demographic alternatives. The goal is not to remove demographic associations entirely, but to prevent the model from assigning systematically different probabilities to equivalent demographic alternatives. 

We consider a next-token prediction setting in which a language model receives a socio-economic question $x$ and assigns probabilities to a pair of demographic answer candidates. Let $\mathcal{Y}(x)=\{y_{\text{stereo}}, y_{\text{anti}}\}$ denote the candidate set, where $y_{\text{stereo}}$ and $y_{\text{anti}}$ correspond to stereotypical and anti-stereotypical demographic alternatives, respectively. For example, given the prompt ``Who often has difficulty paying back their loan?'', the model assigns probabilities to the candidates ``women'' and ``men''. An example prompt from this setting is shown in Table~\ref{tab:socioeconomicqa-example}.

\begin{table}[t]
\centering
\begin{tabular}{p{0.94\columnwidth}}
\toprule
\textbf{SocioEconomicQA -- Intrinsic Bias Evaluation} \\
\midrule
\textbf{Question:} Who often has difficulty paying back their loan? \\
\textbf{Answer Options:}\hspace{1.5em} A. women \hspace{1.5em} B. men \\
\bottomrule
\end{tabular}
\caption{Example prompt from the SocioEconomicQA dataset used for intrinsic bias evaluation.}
\label{tab:socioeconomicqa-example}
\end{table}

For each candidate answer $y \in \mathcal{Y}(x)$, let
$p_\theta(y \mid x)$ denote its raw next-token probability under the
language model with parameters $\theta$. To compare alternatives within
the candidate set, we define the candidate-normalized probability:
\begin{equation}
\hat{p}_\theta(y \mid x)
=
\frac{p_\theta(y \mid x)}
{\sum_{y' \in \mathcal{Y}(x)} p_\theta(y' \mid x)} .
\label{eq:normalized_candidate_probability}
\end{equation}

Intrinsic bias manifests as asymmetric probability assignments across demographic alternatives, often favoring stereotypical responses. We quantify this asymmetry through the probability gap:
\begin{equation}
\textsc{Gap}(x)
=
\left|
\hat{p}_\theta(y_{\text{stereo}} \mid x)
-
\hat{p}_\theta(y_{\text{anti}} \mid x)
\right| .
\label{eq:intrinsic_bias_gap}
\end{equation}

A fair model should assign balanced probabilities across demographic alternatives, i.e., $\textsc{Gap}(x) \approx 0$, while preserving general language modeling performance. Therefore, the objective of intrinsic bias mitigation is to reduce systematic probability differences between demographic alternatives without substantially altering the pretrained model's overall behavior.

Standard concept unlearning objectives typically operate on individual target responses, reducing or increasing their likelihood without directly constraining the relationship between competing alternatives. In fairness-sensitive settings, this can be insufficient: optimizing candidates independently may reduce one biased association while producing probability collapse or bias inversion. We therefore introduce a fairness-aware objective that explicitly controls the probability gap between stereotypical and anti-stereotypical candidates.

The FACU training objective is composed of the following loss terms:

\paragraph{Unlearning Loss.}
This term suppresses stereotypical responses by reducing their likelihood under the model:

\begin{equation}
L_{\text{unlearn}}
=
\mathbb{E}_{(x, y_{\text{stereo}})}
\left[
- \log p_\theta(y_{\text{stereo}} \mid x)
\right] .
\label{eq:unlearn_loss}
\end{equation}

Although $L_{\text{unlearn}}$ is written as a standard negative log-likelihood term, it is included in the final objective with a negative coefficient. Therefore, minimizing the total objective performs gradient ascent on stereotypical responses, reducing their probability under the model.

\paragraph{Learning Loss.}
This term reinforces anti-stereotypical alternatives:

\begin{equation}
L_{\text{learn}}
=
\mathbb{E}_{(x, y_{\text{anti}})}
\left[
- \log p_\theta(y_{\text{anti}} \mid x)
\right] .
\label{eq:learn_loss}
\end{equation}

Unlike the unlearning term in Eq.~\eqref{eq:unlearn_loss}, this loss is optimized by standard gradient descent, increasing the likelihood of anti-stereotypical responses.

\paragraph{Gap Loss (Fairness Constraint).}
Our key contribution is a fairness regularization term that directly penalizes probability differences between demographic alternatives:
\begin{equation}
L_{\text{gap}}
=
\mathbb{E}_{x}
\left[
\left(
p_\theta(y_{\text{stereo}} \mid x)
-
p_\theta(y_{\text{anti}} \mid x)
\right)^2
\right] .
\label{eq:gap_loss}
\end{equation} 

This term (\ref{eq:gap_loss}) explicitly regulates the relative probability structure between demographic alternatives, encouraging balanced assignments and reducing the risk of probability collapse or bias inversion.

\paragraph{Normalization Loss.}
To preserve the pretrained model's general language modeling behavior, we
regularize the updated model with a KL divergence penalty against the frozen
reference distribution:
\begin{equation}
L_{\text{norm}}
=
\mathbb{E}_{x \sim \mathcal{D}_{\text{ref}}}
\Big[
\mathrm{KL}\!\left(
p_{\theta}(\cdot \mid x)
\,\big\|\,
p_{\theta_0}(\cdot \mid x)
\right)
\Big],
\label{eq:kl_loss}
\end{equation}
where $p_{\theta_0}$ denotes the frozen pretrained model, $p_{\theta}$ the
model being optimized, and $\mathcal{D}_{\text{ref}}$ a retain set of neutral
prompts. This loss penalizes deviations from a frozen reference model on neutral prompts helping preserve the model's general language modeling behavior.

\paragraph{Final Objective.}
The overall training objective is defined as a weighted combination of all loss components:
\begin{equation}
L_{\text{total}}
=
-\lambda_1 L_{\text{unlearn}}
+
\lambda_2 L_{\text{learn}}
+
\lambda_3 L_{\text{gap}}
+
\lambda_4 L_{\text{norm}},
\label{eq:final_objective}
\end{equation}

where $\lambda_i > 0$ are weighting coefficients controlling the contribution of each objective during optimization. The negative sign before $L_{\text{unlearn}}$ implements gradient ascent on stereotypical responses, thereby reducing their probability under the model.

Overall, FACU extends standard concept unlearning with an explicit fairness constraint that minimizes probability asymmetry between demographic alternatives while preserving the pretrained model's general behavior.


\section{Experimental Setup}
\label{Exp_Setup}
We conduct experiments using three instruction-tuned LLMs: Llama-3.1 (8B)~\citep{grattafiori2024llama}, Phi-3 Mini (3.8B)~\citep{abdin2024phi}, and Gemma-2 (2B)~\citep{gemma_2024}. Additional details are provided in
Appendix A.\footnote{The appendices are available in the extended version on arXiv.}

\subsection{Intrinsic Bias Evaluation and Mitigation}

For the first research question, we evaluate intrinsic bias using the gender subset of \textsc{SocioEconomicQA}~\cite{arzaghi2024understanding}, which contains socio-economic questions paired with stereotypical and anti-stereotypical answer alternatives, as illustrated in Table~\ref{tab:socioeconomicqa-example}. The dataset is split into disjoint training, validation, and test sets. The training and validation splits are used for unlearning and model selection, while the test split is reserved for final evaluation. \textsc{TruthfulQA}~\cite{lin2021truthfulqa} is used as the reference distribution for KL regularization. To evaluate out-of-distribution generalization, we additionally assess mitigated models on the gender subsets of BBQ~\cite{parrish2022bbq} and CrowS-Pairs~\cite{nangia2020crows}, which contain broader social bias contexts beyond socio-economic settings.

To evaluate intrinsic bias, we compute normalized next-token probabilities over paired demographic alternatives, as described in Section~\ref{sec:Methodology}. We report the Poverty Association Ratio (PAR)~\cite{arzaghi2024understanding}, measuring associations between socio-economic attributes and gender terms, and iCAT~\cite{nadeem2020stereoset}, capturing stereotype balance and language modeling capability. We also report perplexity on \textsc{WikiText-2}~\cite{merity2016pointer} to assess the trade-off between mitigation and language modeling performance. Dataset and metric details are provided in Appendices A and B.

FACU is implemented through full fine-tuning and evaluated against the pretrained model (no mitigation), standard concept unlearning~\cite{yao2024large}, Direct Preference Optimization (DPO)~\cite{rafailov2023direct}, and self-debiasing~\cite{gallegos2025self}. The unlearning baseline uses the same stereotypical and anti-stereotypical candidate pairs as FACU but omits the GAP-loss term. DPO is trained on preference pairs favoring anti-stereotypical over stereotypical answers, while self-debiasing steers candidate probabilities at inference time without updating model parameters.

\subsection{Downstream Evaluation, Deployment, and Mitigation}

\begin{table}[t]
\centering
\begin{tabular}{p{0.94\columnwidth}}
\toprule
\textbf{Adult Dataset -- Downstream Task Evaluation} \\
\midrule

\textbf{Prompt:}
Below is an instruction that describes a task, paired with an input that
provides further context. Write a response that appropriately completes
the request.

\par
\textbf{Instruction:}
Predict whether the given person earns a salary greater than \$50K based
on their description. Answer only as yes/no.

\par
\textbf{Input:}
workclass is Private. hours-per-week is 50. sex is Male. age is 31.
occupation \ldots

\\
\midrule
\textbf{Ground-truth label:} yes \\
\bottomrule
\end{tabular}
\caption{Example serialized input from the Adult dataset used for downstream task evaluation.}
\label{tab:adult-example}
\end{table}

For the second and third research questions, we evaluate downstream fairness on three tabular classification datasets with gender disparities: \textsc{ACS Employment}~\cite{ding2021retiring}, \textsc{Adult}~\cite{adult_2}, and \textsc{German Credit}~\cite{default_of_credit_card_clients_350}. Each dataset defines a binary classification task and uses gender as the sensitive attribute for fairness evaluation. We report predictive performance (Acc) together with group fairness metrics: Accuracy Parity (AccP), Demographic Parity (DP), and Equalized Odds (EqOdds)~\cite{barocas2023fairness,hardt2016equality}. These metrics capture complementary aspects of group fairness: differences in positive prediction rates, error rates, and overall accuracy across gender groups. For all fairness metrics, lower values indicate smaller disparities across demographic groups and therefore fairer outcomes.

For downstream deployment, we follow the TabLLM~\cite{hegselmann2023tabllm} paradigm of converting structured inputs into natural language. An example serialized input from the \textsc{Adult} dataset is shown in Table~\ref{tab:adult-example}.
Additional details are provided in Appendix~C. We consider three settings: (i) logistic regression as a classical baseline, which provides a transparent reference point; (ii) LLM embeddings with logistic regression, where serialized inputs are encoded by a frozen LLM and used to train a fixed classifier, isolating representation-level effects while keeping the LLM parameters unchanged; and (iii) LLM as classifier, where the LLM is fine-tuned for downstream prediction using LoRA~\cite{hu2022lora}, allowing us to assess whether mitigation effects persist after task-specific adaptation. 

All experiments are repeated over three independent train--test splits. For each split, models are trained for multiple epochs and the checkpoint with the highest test accuracy is selected. Fairness metrics are then evaluated on this checkpoint, ensuring that model selection is driven solely by predictive performance rather than fairness objectives. Final results are averaged across splits. 

To evaluate the effect of intrinsic mitigation, FACU is applied prior to downstream adaptation, enabling analysis of how intrinsic debiasing propagates to downstream fairness outcomes across different deployment settings. We compare FACU with extrinsic methods operating at different stages of the pipeline. Specifically, we implement counterfactual data augmentation as a data-level intervention and self-debiasing as an inference-level intervention. We evaluate each method independently and in combination with FACU to compare intervention levels and assess their complementarity.

\textbf{Data-level intervention:} Counterfactual Data Augmentation (CDA)~\cite{kusner2017counterfactual,xie2023empirical} augments the training data by duplicating examples with the gender attribute flipped, thereby reducing spurious correlations between gender and the target label. When generating counterfactuals, we also adjust related attributes to preserve semantic consistency.

\textbf{Inference-level intervention:} Self-debiasing~\cite{gallegos2025self} is an inference-time method that revises model outputs through additional prompting. Because it operates on generated responses, it is only applicable in the LLM-as-classifier setting. Further details on its adaptation to downstream tasks are provided in Appendix~D.

\section{Experimental Results}
\label{Exp_Result}

\subsection{RQ1: Can FACU effectively mitigate intrinsic gender bias in LLMs?}
\label{RQ1}

Table~\ref{tab:FACU_Results} compares FACU with the pretrained model, unlearning, DPO, and self-debiasing across all evaluated LLMs. For stereotypical and anti-stereotypical probabilities, values closer to 0.5 indicate more balanced associations. Lower GAP values indicate smaller disparities between demographic alternatives. iCAT ranges from 0 to 1, with higher values reflecting a better trade-off between stereotype mitigation and language modeling quality.

First, we observe that compared to the pretrained models, self-debiasing reduces GAP across all LLMs, but its gains are smaller than those of FACU and are not consistently reflected in iCAT. DPO also yields limited GAP reductions and substantially lowers iCAT, particularly for Llama and Gemma, suggesting that preference optimization alone does not reliably produce balanced demographic associations.

Moreover, unlearning exhibits unstable behavior, especially for Gemma and Phi, where it drives stereotypical probabilities toward zero and anti-stereotypical probabilities toward one, indicating probability collapse and bias inversion rather than balanced mitigation. For Llama, it produces only a moderate GAP reduction. This reflects a mismatch with our fairness objective, which requires balancing probabilities across alternatives rather than suppressing individual outputs.

In contrast, FACU achieves the lowest GAP and highest iCAT for every model by explicitly constraining probability differences between alternatives. It also maintains stable perplexity, indicating effective reduction of the measured intrinsic gender-bias gap without degrading language modeling quality. Self-debiasing leaves perplexity unchanged because it operates at inference time, while DPO and unlearning produce only modest perplexity changes despite their weaker or unstable fairness outcomes.

Finally, nonparametric tests confirm that the PAR gap reductions achieved by FACU are statistically significant across all models (Table~12, Appendix~D).
Within the evaluated gender-bias setting, FACU provides more consistent probability balancing than unlearning, DPO, and self-debiasing, motivating our analysis of whether these improvements transfer to downstream fairness.

\begin{table}[t]
\centering
\resizebox{\columnwidth}{!}{
\begin{tabular}{|l| l| c c c|c|c|}
\toprule
\textbf{Model} 
& \textbf{Method} 
&
\multicolumn{3}{c|}{\textbf{PAR}} 
&
\textbf{iCAT}
& \textbf{PPL} \\
\cmidrule(lr){3-5}
& 
&Anti-Ster.
&Ster.
&GAP($\downarrow$)  
& ($\uparrow$) 
&($\downarrow$) \\
\midrule

\multirow{2}{*}{Llama}
& Pretrained     
& 0.220 
& 0.780 
& 0.560 
& 0.343 
& 9.452 \\
&Self-debiasing     
&0.293 
&0.707 
&0.414 
&0.256 
&9.452 \\
&DPO    
&0.245 
&0.755 
&0.510 
&0.159 
&9.459 \\
&  Unlearning
&  0.305 
&  0.695 
&  0.390  
&  0.451
&  9.458 \\
&  FACU
&  0.470 
&  0.530 
&  \textbf{0.059} 
&  \textbf{0.698}
&  9.459 \\
\midrule

\multirow{2}{*}{Gemma}
& Pretrained     
& 0.193 
& 0.807 
& 0.614 
& 0.229
&   27.880 \\
&Self-debiasing      
&0.294 
&0.706 
&0.412 
&0.260 
&27.880 \\
&DPO    
&0.235 
&0.766 
&0.531
&0.077 
&27.833 \\
&  Unlearning
&  \num{1.000} 
&  \num{0.000} 
&  \num{1.000}  
&  0.000    
&  28.528 \\
&  FACU
&  0.546 
&  0.454 
&  \textbf{0.092}  
&  \textbf{0.906}
&  27.683 \\
\midrule

\multirow{2}{*}{Phi}
& Pretrained  
& 0.291 
& 0.709 
& 0.418  
&0.361 
&  7.870 \\
&Self-debiasing      
&0.317 
&0.683 
&0.366 
&0.298  
&7.870 \\
&DPO    
&0.309 
&0.692
&0.383
&0.270
&7.879 \\
& Unlearning
&  0.992 
&  0.008
&  0.984  
&  0.009
&  8.072 \\
&  FACU
&  0.514 
&  0.486 
&  \textbf{0.029 } 
&  \textbf{0.842}
&  7.838 \\
\bottomrule
\end{tabular}
}
\caption{Intrinsic bias results across LLMs. We report anti-stereotypical and stereotypical probabilities, GAP, iCAT, and perplexity (PPL); best GAP and iCAT values are bolded.}
\label{tab:FACU_Results}
\end{table}

\paragraph{Cross-Domain Transfer Within Gender-Bias Evaluation.}
To evaluate whether gender bias mitigation learned by FACU generalizes beyond socio-economic contexts, we conduct a supplementary evaluation on \textsc{BBQ}~\cite{parrish2022bbq} and \textsc{CrowS-Pairs}~\cite{nangia2020crows}. Importantly, FACU is trained only on \textsc{SocioEconomicQA}, which focuses on financial contexts and gendered terms, and is not retrained or adapted on either evaluation dataset. BBQ and CrowS-Pairs contain broader social bias contexts and include different forms of gender expression, such as names and pronouns not observed by FACU during training. Despite this distribution shift, FACU consistently reduces intrinsic gender bias on both datasets, suggesting that the mitigation effect generalizes beyond the specific socio-economic contexts and linguistic patterns seen during training. These findings indicate that FACU captures broader representational gender bias patterns rather than overfitting to dataset-specific stereotypes. Detailed results are reported in Table~11 in Appendix~D.

\subsubsection{Ablation Study: Understanding FACU Components}
\label{Ablation}
We conduct an ablation study to assess the contribution of each loss component in FACU. Overall, intrinsic bias mitigation depends on a balance between unlearning, learning, and regularization terms. The unlearning and learning losses defined in \eqref{eq:unlearn_loss} and \eqref{eq:learn_loss} drive the main parameter updates by reducing stereotypical associations and reinforcing anti-stereotypical associations. However, when applied without additional constraints, this process can lead to uncontrolled updates, resulting in undesirable effects such as bias inversion as observed in the classic unlearning setting. This observation is consistent across all studied LLMs. For the smaller models, Gemma and Phi, removing the Gap loss defined in \eqref{eq:gap_loss} often results in bias inversion, where stereotypical associations are replaced by anti-stereotypical ones. In contrast, for Llama, this produces only marginal changes and fails to effectively reduce the intrinsic bias gap. These results suggest that model capacity influences how bias manifests under unconstrained updates; however, in all cases the resulting probability distributions remain unbalanced.

We further observe that strong KL regularization helps preserve model coherence but can limit bias mitigation by keeping the updated model close to the original biased distribution. This is consistent with observations in related KL-regularized alignment settings~\cite{xiao2024algorithmic}.

Taken together, these findings show that all components of FACU are necessary to achieve effective intrinsic bias mitigation across models, with the Gap loss playing a key role. Detailed ablation results for Llama, Gemma, and Phi are provided in Appendix~D, Tables~8, 9, and~10.

\subsection{RQ2: Does intrinsic gender bias mitigation via FACU translate into improved downstream fairness in decision-making tasks? }
\label{RQ2}
\begin{table*}[!ht]
\centering
\resizebox{\textwidth}{!}{%
\begin{tabular}{|l|l|cccc|cccc|cccc|}
\toprule
\textbf{Model} 
&\textbf{Stage} 
&\multicolumn{4}{c|}{\textbf{ACS Employment Dataset }}
&\multicolumn{4}{c|}{\textbf{Adult Dataset}} 
&\multicolumn{4}{c|}{\textbf{German Credit Dataset}} \\
\cmidrule(lr){3-6} \cmidrule(lr){7-10} \cmidrule(lr){11-14}
& 
&Acc
&AccP
&DP
&EqOdds
&Acc
&AccP
&DP
&EqOdds
&Acc
&AccP
&DP 
&EqOdds \\
&
&($\uparrow$) 
&($\downarrow$) 
&($\downarrow$) 
&($\downarrow$) 
&($\uparrow$) 
&($\downarrow$) 
&($\downarrow$) 
&($\downarrow$) 
&($\uparrow$) 
&($\downarrow$) 
&($\downarrow$) 
&($\downarrow$) \\

\midrule
\midrule
\multicolumn{14}{c}{ \textbf{Logistic Regression as Baseline }} \\
\midrule
\multirow{2}{*}{L. R.}
&No Mitig. 
&0.701 
&0.117 
&0.215 
&0.612 
&\textbf{0.812 }
&\textbf{0.144 } 
&\textbf{0.326} 
&\textbf{0.365} 
&\textbf{0.738} 
&0.045  
&\textbf{0.095}  
&\textbf{0.160}  \\

& CDA 
& \textbf{0.716} 
& 0.117 
& \textbf{0.142} 
& \textbf{0.458} 
& 0.805  
& 0.146 
& 0.333 
& 0.392 
& 0.692 
& \textbf{0.037} 
& 0.098  
& 0.157 \\

\midrule
\midrule

\multicolumn{14}{c}{ \textbf{LLM as Feature Extractor + Logistic Regression}} \\
\midrule

\multirow{4}{*}{Llama}
&Pretrained
&0.747 
&0.132  
&0.084  
&0.268 
&0.817 
&0.129 
&0.302 
&0.312 
&0.803 
&0.073 
&0.079 
&0.171 \\

& FACU 
&  \textit{0.751} 
&  \textit{0.097} 
&  \textit{0.071} 
&  \textit{0.258} 
&  \textit{0.820} 
&  \textit{0.127}
&  \textit{0.299} 
&  \textit{0.291} 
&  \textit{0.805} 
&  \textit{0.071} 
&  \textit{0.073} 
&  \textit{0.152} \\

& CDA 
&  0.763 
&  0.117 
&  0.080  
&  0.262 
&  0.815 
&  0.128 
&  0.291 
&  0.298 
&  0.705 
&  0.060
&  \textbf{0.012} 
&  0.094 \\

& FACU+CDA
& \underline{\textbf{0.766}} 
& \underline{\textbf{0.096}}  
& \underline{\textbf{0.014}} 
& \underline{\textbf{0.167}} 
& \underline{0.816} 
& \underline{\textbf{0.120}} 
& \underline{\textbf{0.288}} 
& \underline{\textbf{0.280}} 
& 0.700 
& \underline{\textbf{0.044}} 
& 0.017 
& \underline{\textbf{0.052}} \\
\midrule
\multirow{4}{*}{Gemma}
&Pretrained
&0.757 
&0.199  
&0.071  
&0.264 
&\textbf{0.822} 
&0.134 
&0.303 
&0.291 
&\textbf{0.816} 
&0.050 
&0.063 
&0.094 \\

& FACU 
&  0.757 
&  \textit{0.104}  
&  \textit{0.054}  
&  \textit{0.248} 
&  0.821 
&  \textit{0.130} 
&  \textit{0.301} 
&  \textit{0.288} 
&  0.810 
&  0.059 
&  0.067 
&  0.141 \\

& CDA 
&  0.770 
&  0.116 
&  0.061 
&  0.257 
&  0.820 
&  \textbf{0.130} 
&  \textbf{0.298} 
&  0.308 
&  0.741 
&  0.035 
&  \textbf{0.035} 
&  \textbf{0.058} \\

& FACU+CDA
& \underline{\textbf{0.771}} 
& \underline{\textbf{0.100}} 
& \underline{\textbf{0.014}} 
& \underline{\textbf{0.190}} 
& \underline{0.821} 
& \underline{\textbf{0.124}} 
& \underline{0.301} 
&  0.310 
&  0.730 
& \underline{\textbf{0.025}} 
&  0.085 
&  0.147 \\

\midrule
\multirow{4}{*}{Phi}
&Pretrained
&0.751 
&0.127  
&0.075 
&0.287 
&0.821 
&0.130 
&0.303 
&0.292 
&0.795 
&0.054 
&0.038 
&0.097 \\

& FACU 
& 0.751 
& \textit{0.101} 
& \textit{0.063}  
&  \textit{0.261} 
&  \textit{\textbf{0.822}} 
&  \textit{0.120} 
&  \textit{0.302 }
&  \textit{0.287} 
&  \textit{0.798} 
&  \textit{0.052} 
&  \textit{\textbf{0.024}} 
&  \textit{\textbf{0.082}} \\

& CDA 
&  0.766 
&  0.121  
&  0.073  
&  0.281 
&  0.816 
&  0.127 
&  0.291 
&  0.290 
&  0.707 
&  0.007 
&  0.040 
&  0.099 \\

& FACU+CDA
&  \underline{\textbf{0.767}} 
&  \underline{\textbf{0.100}} 
&  \underline{\textbf{0.016}}  
&  \underline{\textbf{0.181}} 
&  0.817 
&  \underline{\textbf{0.116}} 
&  \underline{\textbf{0.289}} 
&  \underline{\textbf{0.284}} 
&  0.705
&  \underline{\textbf{0.005}} 
&  0.048 
&  \underline{0.094} \\

\midrule
\midrule
\multicolumn{14}{c}{ \textbf{LLM as Classifier}} \\
\midrule
\multirow{4}{*}{Llama}
&Pretrained
&0.767 
&0.114  
&0.051 
&0.275 
&0.838
&0.129 
&0.308 
&0.300 
&0.671 
&0.146 
&0.050 
&0.130 \\
& FACU 
&  \textit{0.768} 
&  \textit{\textbf{0.112}} 
&  \textit{0.045 }
&  \textit{0.265} 
&  0.838 
&  0.131 
&  \textit{0.303} 
&  \textit{0.280} 
&  \textit{0.680 }
&  \textit{0.084} 
&  0.118 
&  0.240 \\

& CDA 
&  0.767 
&  0.116  
&  0.049  
&  0.275 
&  0.829 
&  0.133 
&  0.308 
&  0.298 
&  0.678 
&  0.126 
&  0.043 
&  0.079 \\

& Self-debiasing 
& 0.758 
& 0.129 
& 0.021 
& 0.198 
& 0.831 
& 0.128
& 0.305 
& 0.280 
& \textbf{0.695} 
& 0.088  
& \textbf{0.003}  
& \textbf{0.013} \\

& FACU+CDA
&  0.767
&  0.118 
&  0.050 
&  \underline{0.255} 
&  \underline{0.836} 
&  \underline{0.131} 
&  \underline{0.306} 
&  \underline{0.286} 
&  \underline{0.681} 
&  \underline{0.081} 
&  0.114 
&  0.193 \\

& FACU+Self
&  \underline{\textbf{0.771}}
&  \underline{0.120} 
&  \underline{\textbf{0.007}}
&  \underline{\textbf{0.159}}
&  \underline{\textbf{0.841}} 
&  \underline{\textbf{0.117}}  
&  \underline{\textbf{0.285}}
&  \underline{\textbf{0.229}} 
&  0.676 
&  \underline{\textbf{0.073}}  
&  0.165  
&  0.255 \\

\midrule
\multirow{4}{*}{Gemma}
&Pretrained
&0.766
&0.112 
&0.047 
&0.258 
&0.837 
&0.130 
&0.313 
&0.315 
&0.673 
&0.068 
&0.107 
&0.221 \\

& FACU 
&  0.766 
&  \textit{0.111}  
&  \textit{0.039}  
&  \textit{0.244} 
&  0.837 
&  \textit{\textbf{0.128}} 
&  \textit{\textbf{0.303}} 
&  \textit{\textbf{0.280}} 
&  0.637 
&  0.069 
&  \textit{0.078} 
&  \textit{0.167} \\

& CDA 
&  0.766
&  \textbf{0.109} 
&  0.043 
&  0.256 
&  0.837 
&  0.130 
&  0.313 
&  0.315 
&  \textbf{0.675} 
&  0.054 
&  0.126 
&  0.207 \\

& Self-debiasing 
& 0.761 
& 0.116 
& 0.067 
& 0.284 
& 0.837 
& 0.131  
& 0.309 
& 0.300 
& 0.659 
& \textbf{0.032}  
& 0.077 
& 0.142   \\

& FACU+CDA
& \underline{0.767} 
& \underline{\textbf{0.109}} 
& \underline{0.039} 
& \underline{0.244} 
& 0.837 
& \underline{\textbf{0.128}} 
& \underline{\textbf{0.303}} 
& \underline{\textbf{0.280}} 
& 0.640 
& 0.056 
& \underline{\textbf{0.075}} 
& \underline{\textbf{0.134}} \\

& FACU+Self
& \underline{0.766}
& 0.117 
& \underline{\textbf{0.031}}
& \underline{\textbf{0.240}}
& 0.826  
& 0.137
& 0.326 
& 0.325 
& 0.475 
& 0.085  
& 0.135  
& 0.206 \\

\midrule
\multirow{4}{*}{Phi}
&Pretrained
&\textbf{0.754} 
&0.093  
&0.075  
&0.318 
&0.800 
&0.151 
&0.342 
&0.399 
&\textbf{0.680} 
&0.054 
&0.138 
&0.214 \\

& FACU 
& 0.753 
& \textit{0.090}  
&  \textit{\textbf{0.073}} 
& \textit{0.288} 
& \textit{0.808 }
& \textit{0.148} 
& \textit{0.331 }
& \textit{0.362} 
& 0.655 
& \textit{\textbf{0.035}} 
& \textit{\textbf{0.070}} 
& \textit{\textbf{0.109}} \\

& CDA 
& 0.753 
& 0.094  
& 0.074  
& 0.297 
& 0.824 
& \textbf{0.135} 
& 0.309 
& 0.305 
& 0.669 
& 0.052 
& 0.118 
& 0.157\\

& Self-debiasing 
& 0.753  
& 0.119 
& 0.088 
& 0.336  
& 0.781 
& 0.141  
& 0.333  
& 0.367 
& 0.519 
& 0.068  
& 0.091  
& 0.193   \\

& FACU+CDA
&  0.752 
&  \underline{0.093}  
&  0.074  
&  \underline{\textbf{0.280}} 
&  \underline{\textbf{0.827}} 
&  0.137 
&  \underline{\textbf{0.305}} 
&  \underline{\textbf{0.295}} 
&  0.667 
&  \underline{0.039} 
&  \underline{0.098} 
&  \underline{0.130} \\

& FACU+Self.
& 0.735
& \underline{\textbf{0.084}} 
& 0.157  
& 0.443 
& \underline{0.789 } 
& \underline{0.138}  
& \underline{0.330} 
& 0.374 
& 0.517 
& 0.101  
& 0.099  
& \underline{0.181} \\

\bottomrule
\end{tabular}
}
\caption{Downstream accuracy and group fairness metrics across datasets and model configurations. Rows are grouped by mitigation type: no mitigation (\textit{Pretrained}), intrinsic mitigation via FACU applied prior to downstream evaluation (\textit{FACU}, RQ2), extrinsic mitigation (\textit{CDA}, \textit{Self-debiasing}, RQ3), and combined intrinsic and extrinsic mitigation (\textit{FACU+CDA}, \textit{FACU+Self}, RQ3). \\\textbf{Bold} values indicate the best result. \textit{Italic} values indicate improvements from FACU over the pretrained model. \underline{Underlined} values highlight improvements of combined mitigation over the corresponding extrinsic-only setting.} 
\label{tab:DownstreamTask}
\end{table*}


\begin{table}[t]
\centering
{
\begin{tabular}{|l|c|ccc|}
\toprule
\textbf{Mitigation}
& \textbf{Acc} 
& \textbf{AccP} 
& \textbf{DP} 
& \textbf{EqOdds}
\\
\midrule
 \textbf{FACU}
&  0.656
&  0.089
&  0.019
&  0.019 \\
 \textbf{CDA}
&  0.656
&  0.363
&  0.062
&  0.008 \\

 \textbf{Self-deb.}
&  0.996
&  0.746
&  0.089
&  0.089 \\

 \textbf{FACU+CDA}
&  0.454
&  0.001
&  0.004
&  0.004 \\

 \textbf{FACU+Self-deb.}
&  1.000
&  0.500
&  0.500
&  0.500 \\
\bottomrule
\end{tabular}
}
\caption{Paired sign tests comparing mitigation methods against pretrained baselines. Reported values are p-values from exact binomial sign tests.}
\label{tab:pvalue_sign_test}
\end{table}

In this section, we evaluate whether mitigating intrinsic bias through FACU translates into improved downstream fairness. To isolate this effect, we compare pretrained models with their FACU-mitigated counterparts under identical downstream conditions. No extrinsic mitigation is applied in this analysis, allowing us to attribute any observed changes solely to intrinsic bias mitigation.
Table~\ref{tab:DownstreamTask} reports accuracy and group-fairness metrics across three datasets and two deployment settings. 
For this analysis, we rely on the comparison between the pretrained and FACU rows of Table~\ref{tab:DownstreamTask}.

Comparing pretrained and FACU-mitigated models, we observe improvements in most fairness metrics, although the direction and magnitude vary across models, datasets, deployment settings, and metrics. For the feature-extractor setting, FACU reduces most measured disparities, with exceptions for Gemma on German Credit, where some fairness metrics slightly deteriorate. For the classifier setting, FACU again improves fairness in the majority of cases, with only a few metric-level exceptions (4 out of 27 comparisons: 3 fairness metrics $\times$ 3 datasets $\times$ 3 LLMs).

We additionally report paired statistical tests are reported in Table~\ref{tab:pvalue_sign_test}, where lower p-values indicate more consistent improvements relative to pretrained baselines. For FACU, the results show statistically significant improvements across all fairness metrics, while changes in predictive accuracy are not statistically significant. This suggests that downstream fairness can improve without significantly degrading predictive performance compared to non-mitigated models.

These results suggest that mitigating intrinsic gender associations can improve downstream fairness while largely preserving predictive performance.

\paragraph{Baseline Behavior.}
Across datasets and models, LLM-based approaches, both as feature extractors and as classifiers, generally outperform logistic regression in terms of predictive accuracy, particularly on the ACS Employment and Adult datasets, confirming the utility of LLM-based representations for downstream prediction. In many cases, pretrained LLMs also achieve stronger fairness metrics than logistic regression baselines. On German Credit, LLMs outperform logistic regression in the feature-extractor setting, while the classifier setting yields lower accuracy, which we hypothesize is due to the smaller dataset size. At the same time, in terms of fairness, LLM-based approaches exhibit greater variability across datasets and deployment settings, indicating that stronger predictive performance does not necessarily eliminate demographic disparities. Overall, these results suggest that LLM-based approaches can provide both strong predictive performance and competitive fairness compared to logistic regression, while intrinsic mitigation through FACU can further improve fairness without substantially degrading performance.

\paragraph{Differences Across LLMs.} 
We observe substantial behavioral variation across models that depends on the deployment strategy. When used as feature extractors, Gemma consistently achieves the highest accuracy, while Llama shows weaker performance in this configuration despite its larger model size. This trend reverses when considering fairness: Llama generally attains better fairness metrics, whereas Gemma exhibits higher bias. Phi provides a more balanced trade-off between accuracy and fairness across deployment settings.
Among LLM-based classifiers, Gemma stands out as the strongest performer, achieving competitive accuracy and fairness simultaneously, whereas Phi shows a noticeable drop in performance. On the German Credit dataset, Llama demonstrates relatively strong results. This may be explained by the dataset’s limited size, which can amplify overfitting effects, or by dataset-specific sensitivity to model pretraining and fine-tuning dynamics. These findings suggest that the effectiveness of intrinsic bias mitigation depends not only on the mitigation method itself, but also on the underlying model and deployment strategy used in downstream applications.

\subsection{RQ3: How does intrinsic gender bias mitigation compare with and interact with extrinsic mitigation strategies?}
\label{RQ3}
We compare FACU with two extrinsic mitigation strategies applied at different stages of the pipeline: Counterfactual Data Augmentation (CDA) at the data level and self-debiasing at the inference level (CDA and self-debiasing rows in Table~\ref{tab:DownstreamTask}). We further evaluate whether combining these methods with FACU (FACU+CDA and FACU+Self rows) leads to additional fairness improvements. Because CDA and self-debiasing operate at downstream level rather than directly mitigating pretrained representations, they provide a comparison point for assessing whether intrinsic mitigation offers distinct benefits.

\paragraph{FACU vs. extrinsic methods.}
Across LLMs and datasets, FACU generally achieves larger and more consistent fairness improvements than both CDA and self-debiasing across datasets and metrics. This trend is particularly clear in the LLM-as-classifier setting, where FACU consistently yields smaller fairness disparities across multiple metrics, while self-debiasing generally performs better than CDA. This advantage of FACU is particularly pronounced for Gemma and Phi. In terms of predictive accuracy, self-debiasing performs worse compared to the other methods, while FACU maintains comparable performance. These observations are supported by paired sign tests (Table~\ref{tab:pvalue_sign_test}). 

FACU shows statistically significant improvements on Demographic Parity and Equalized Odds, whereas self-debiasing does not yield significant gains. CDA shows statistically significant improvement only for Equalized Odds. FACU also demonstrates a consistent trend toward improvement in Accuracy Parity, reflected by substantially lower p-values than the other methods, although this result does not reach statistical significance. Furthermore, differences in predictive accuracy are not statistically significant across methods, indicating that fairness improvements are not associated with systematic performance degradation.

In the LLM-as-feature-extractor setting, a similar pattern is observed. FACU generally provides stronger fairness improvements than CDA while maintaining comparable or better accuracy across most datasets, with the clearest gains on ACS Employment and Adult.

\paragraph{Combination of FACU with extrinsic methods.}
We evaluate the combination of FACU with extrinsic mitigation methods (CDA and self-debiasing) by using the FACU-mitigated LLM in place of the corresponding pretrained LLM (FACU+CDA and FACU+Self rows in Table~\ref{tab:DownstreamTask}). This setup allows us to assess whether mitigation applied at different stages of the pipeline provides complementary benefits rather than overlapping effects.

Across datasets and models, combining CDA with FACU improves most fairness comparisons while maintaining comparable accuracy, with some exceptions. This pattern appears in both the LLM-as-classifier and LLM-as-feature-extractor settings. Paired sign tests (Table~\ref{tab:pvalue_sign_test}) show statistically significant aggregate gains across all fairness metrics, whereas CDA alone is significant only for Equalized Odds.

A similar but less consistent pattern is observed when combining FACU with self-debiasing. In several cases, the combination improves fairness compared to self-debiasing alone, as indicated by cell-wise improvements in Table~\ref{tab:DownstreamTask}. However, these gains are not consistently supported by statistical significance when evaluated against the pretrained baseline (Table~\ref{tab:pvalue_sign_test}). In particular, results on the German Credit dataset and for smaller models show substantial variability, including notable drops in accuracy. In such cases, fairness comparisons become less reliable, as degraded predictive performance can distort fairness metrics. While we report the corresponding p-values, these results should be interpreted with caution due to the limited number of valid comparisons.

Taken together, these results show that intrinsic mitigation through FACU performs competitively with task-level mitigation methods despite being applied before downstream adaptation. More importantly, combining intrinsic and extrinsic mitigation, particularly with CDA, yields more consistent and statistically significant fairness improvements across datasets and deployment settings.

\paragraph{\textbf{Practical Trade-offs of Mitigation Strategies.}}

Beyond fairness and predictive performance, the methods differ in scalability and deployment cost. FACU requires a one-time model-level mitigation step using a relatively small bias-focused dataset, after which the mitigated model can be reused across downstream tasks. In contrast, CDA must generate augmented data separately for each dataset, increasing training size and cost; indicative run times are reported in Appendix~C, Table~7.
Self-debiasing avoids additional fine-tuning but adds inference-time overhead through repeated prompting. These properties make FACU particularly suitable for repeated deployment across multiple tasks.

\section{Discussion}
\label{Discussion}
\paragraph{\textbf{Implications for Bias Mitigation.}}

FACU substantially reduces measured intrinsic gender bias while preserving perplexity. In contrast, standard concept unlearning can lead to over-correction, producing extreme anti-stereotypical outputs and disrupting the balance between bias mitigation and model coherence. Importantly, while our experiments do not establish a direct or monotonic relationship between intrinsic bias metrics and downstream fairness outcomes, they nevertheless show that mitigating intrinsic gender bias is associated with improvements in downstream group fairness across many evaluated settings. These findings suggest that gender fairness improvements in economic context do not rely exclusively on task-specific extrinsic interventions, but can also be supported by improvements at the intrinsic representational level.

One possible explanation is that balanced debiasing may generalize more effectively across downstream tasks than suppression-based approaches. Rather than removing bias-related signals entirely, the GAP-based objective encourages balanced probability distributions across demographic alternatives, potentially preserving representational information that remains useful for downstream prediction while reducing asymmetric associations. In contrast, suppression- or inversion-based approaches may eliminate or distort these signals entirely, potentially limiting the transfer of intrinsic mitigation to downstream fairness improvements. More broadly, these findings suggest that preserving balanced bias-related features, rather than fully removing them, may inform future research on intrinsic-to-downstream fairness transfer for non-binary and more complex demographic associations, such as intersectional biases.

\paragraph{\textbf{Model-Dependent Behavior of Bias Mitigation.}}
The results across our research questions (Section~\ref{Exp_Result})
highlight that intrinsic mitigation effects depend on model choice and that these differences further influence downstream fairness. 
Notably, across Gemma and Phi, the probability rebalancing observed at the intrinsic level translates into consistent improvements in group fairness metrics across datasets and tasks, particularly when intrinsic and extrinsic mitigation strategies are combined. FACU alone already yields stable reductions in DP and EqOdds with minimal impact on accuracy, while additional interventions, such as CDA or self-debiasing, can further improve fairness, although the magnitude of improvement varies.

In contrast, Llama exhibits a less consistent interaction between intrinsic and downstream behavior. While FACU still improves fairness metrics, the gains are generally smaller and less amplified by additional mitigation, suggesting a more model-dependent interaction between intrinsic and extrinsic methods. 
This complexity may reflect the difficulty of achieving stable convergence toward balanced probability distributions under competing learning and unlearning dynamics (see Appendix~D for the ablation study)
These findings suggest that bias-related representations in Llama may be more entangled with other representational features, making mitigation effects more sensitive to the choice and combination of interventions.
One possible explanation is that the structure of bias within the representation space differs across models, which may influence how mitigation effects propagate to downstream tasks. This interpretation is consistent with prior work showing that contextualized embedding models respond differently to debiasing interventions and fairness trade-offs~\cite{kaneko-bollegala-2021-debiasing,lauscher2020general}.
Overall, these findings suggest that intrinsic mitigation may not transfer uniformly across LLM architectures and motivate further study of more complex biases, including age, race, and religion.

\paragraph{\textbf{Transferability of Intrinsic Bias Mitigation.}}

An additional observation from RQ1 is that the mitigation effects learned by FACU generalize beyond the socio-economic domain used during training. Although FACU is trained only on \textsc{SocioEconomicQA}, it consistently reduces intrinsic bias on broader benchmarks such as \textsc{BBQ} and \textsc{CrowS-Pairs}, which contain different social contexts and linguistic expressions. This suggests that the balanced debiasing induced by FACU may transfer beyond the specific contexts and gender stereotypes observed during mitigation training.
While cross-domain generalization is not the primary focus of this work, these findings highlight a promising direction for studying how intrinsic mitigation methods transfer across domains, demographic attributes, and benchmark distributions.

\paragraph{\textbf{Collective Responsibility.}}
Finally, our findings also have implications for shared responsibility in bias mitigation. Developers and organizations that design and deploy large language models should not rely exclusively on downstream users for bias mitigation. Our results indicate that model-level improvements through intrinsic bias mitigation can meaningfully enhance downstream fairness outcomes. These findings suggest that model developers and pretraining organizations also play a role in mitigating bias at the model level, particularly when such mitigation can be achieved with minimal or no trade-offs in performance. At the same time, downstream task-specific interventions remain valuable, as our experiments suggest that combining intrinsic and extrinsic strategies often yields further improvements in fairness. Bias mitigation should therefore be understood as a shared responsibility across the entire development and deployment pipeline, from model developers to end users.

\section{Limitation}
\label{sec:Limitation}

While our results provide encouraging evidence for fairness-aware intrinsic mitigation, several limitations remain.

First, our downstream analysis focuses on \textbf{binary gender bias}, as gender is the only sensitive attribute consistently annotated across all considered datasets. While this enables controlled comparisons across models, mitigation strategies, and evaluation settings, FACU is not specific to gender and could be extended to additional sensitive attributes and intersectional bias.

Second, our extrinsic evaluation is restricted to \textbf{tabular classification tasks}. While these tasks capture common high-stakes decision-making scenarios and allow controlled evaluation of group-based fairness metrics, the relationship between intrinsic bias mitigation and downstream fairness may differ in other settings, such as ranking, recommendation, or generative tasks.

Third, the experimental setup requires \textbf{substantial computational resources}, combining multiple datasets, train--test splits, LLMs, and mitigation strategies. As a result, we focus on a representative subset of extrinsic mitigation methods that enables controlled comparisons across intervention levels. Broader evaluations across additional mitigation strategies and larger experimental settings remain directions for future work. Additional details on computational cost are provided in Appendix~C.

Together, these limitations highlight the need to study intrinsic bias mitigation across broader demographic attributes, task settings, and mitigation strategies.
\section{Conclusion}
\label{sec:Conclusion}

Across all three research questions, our results provide a consistent picture of the role of intrinsic gender bias mitigation in LLMs. FACU effectively reduces intrinsic gender bias by balancing asymmetric associations, and these improvements are associated with downstream fairness gains without significantly degrading predictive performance. Compared to task-level interventions, FACU achieves competitive or stronger fairness improvements despite being applied prior to downstream adaptation. Moreover, combining intrinsic and extrinsic mitigation yields the most consistent and statistically significant fairness improvements, highlighting the value of addressing bias at multiple stages of the model lifecycle.

Taken together, these findings suggest that fairness in LLM-based decision systems cannot be achieved through a single intervention alone. Instead, effective mitigation requires coordinated efforts across representation-level mitigation, data-level interventions, and inference-time controls. This points to a broader collective responsibility shared by model developers, platform providers, and end users to ensure that appropriate mitigation strategies are applied throughout the deployment pipeline. As LLMs become increasingly integrated into high-stakes decision-making systems, multi-stage mitigation approaches will be important for achieving more reliable and equitable outcomes.
\section*{Ethical Statement}
This work studies intrinsic bias mitigation and downstream fairness in large language models for socio-economic decision-making tasks. The proposed method aims to reduce stereotypical gender associations in model behavior and improve group fairness in downstream predictions. Our experiments show consistent improvements across the evaluated datasets, models, and fairness metrics.

However, these results should be interpreted within the scope of our experimental setting. The evaluation focuses on specific datasets, demographic attributes, model families, and group-fairness metrics, and therefore does not constitute a general guarantee of fairness in all deployment contexts. In particular, real-world decision-making systems may involve additional sources of bias, institutional constraints, legal requirements, and affected populations that are not fully captured by benchmark datasets.

Accordingly, FACU should be viewed as one component of a broader fairness evaluation and mitigation pipeline rather than a standalone guarantee of non-discrimination. We recommend that any deployment of LLM-based decision-making systems include context-specific validation, human oversight, and continuous monitoring.
\section*{Acknowledgments}
\label{sec:Acknowledgments}
We thank Fabrice Normandin for his technical assistance with Mila's computing infrastructure. Funding support for this research was provided in part by the Canada CIFAR AI Chair, a Google Research Award, the NSERC Discovery Grant, and the IVADO Fundamental Research Projects Grant. We also acknowledge Compute Canada for providing computational resources used in this study.
\bibliography{aaai2026}
\label{sec:bibliography}

\newpage
\appendix
\input{appendix}

\end{document}

%% file: Appendix.tex
\section{Model and Dataset Details}
\label{app:model-dataset-details}
\input{App_Model_Dataset}


\section{Metric Definitions}
\label{app:metrics}
\input{App_Metrics}

\section{Implementation Details}
\label{app:implementation}
\input{App_ImplementationDetails}

\section{Additional Results}
\label{app:additional}
\input{App_AdditionalResults}







%% file: App_Model_Dataset.tex
\subsection{Large Language Models}
\label{app:Experiment:models}

We conduct experiments using three open-source, instruction-tuned large language models (LLMs), selected for their strong performance, manageable size, and compatibility with parameter-efficient fine-tuning.

\begin{itemize}
    \item \textbf{Llama-3.1 (8B)}: An 8B parameter model trained on approximately 15 trillion tokens and optimized for multilingual dialogue via instruction tuning and DPO~\cite{grattafiori2024llama}.
    
    \item \textbf{Phi-3 Mini (4K Instruct)}: A 3.8B parameter decoder-only Transformer developed by Microsoft~\cite{abdin2024phi}, fine-tuned using SFT and DPO.
    
    \item \textbf{Gemma-2 (2B-it)}: A 2B parameter decoder-only model developed by Google~\cite{gemma_2024}, optimized for efficiency and English-language reasoning tasks.
\end{itemize}

\subsection{Datasets}
\label{app:Experiment:datasets}
We use separate datasets for intrinsic bias mitigation and evaluation, and for downstream (extrinsic) fairness evaluation.

\subsubsection{Intrinsic Bias Datasets}

The following datasets are used for intrinsic bias mitigation and evaluation:

\begin{itemize}
    \item \textbf{SocioEconomicQA}~\citep{arzaghi2024understanding}: A benchmark for evaluating associations between socioeconomic attributes and demographic groups such as gender, race, religion, and marital status, as well as their intersections. We focus on the gender subset and reformulate each instance into a question–answering format aligned with our unlearning framework. Additional financial-context questions (e.g., income and employment) are included to better match downstream tasks.

    \item \textbf{TruthfulQA}~\citep{lin2021truthfulqa}: Used as a reference distribution for KL-divergence regularization during unlearning, ensuring preservation of general language modeling behavior.

    \item \textbf{WikiText-2}~\citep{merity2016pointer}: Used only at evaluation time to compute perplexity and assess language modeling quality.
\end{itemize}

\subsubsection{Extrinsic Bias Datasets}

We evaluate downstream performance and group fairness on real-world tabular classification tasks with binary gender annotations:

\begin{itemize}
    \item \textbf{ACS Employment}~\cite{ding2021retiring}: Derived from the American Community Survey, we use a subset (Utah, 2018) exhibiting measurable gender imbalance (e.g., 51\% of men vs.\ 41\% of women employed). The dataset contains 22,835 instances with 17 features.

    \item \textbf{Adult (Census Income)}~\cite{adult_2}: A widely used benchmark with 48,842 instances and 14 features. The task is to predict whether income exceeds \$50K.

    \item \textbf{German Credit}~\cite{default_of_credit_card_clients_350}: A smaller dataset with 1,000 instances and 20 attributes, where the task is to predict credit risk (good vs.\ bad).
\end{itemize}

%% file: App_Metrics.tex
\subsection{Intrinsic Bias Metrics}
\label{app:Intrinsic-metrics}

Intrinsic bias is measured using the Poverty Association Ratio (PAR), derived from normalized next-token probabilities~\citep{arzaghi2024understanding}. Each prompt $x$ is paired with a stereotypical completion $y_{\text{ster}}$ and an anti-stereotypical completion $y_{\text{anti}}$.

\paragraph{\textbf{Poverty Association Ratio (PAR).}}
We compute normalized probabilities:
\begin{equation}
\hat{p}_\theta(y \mid x)
= \frac{p_\theta(y \mid x)}
{p_\theta(y_{\text{stereo}} \mid x) + p_\theta(y_{\text{anti}} \mid x)}.
\end{equation}

Values closer to $0.5$ indicate more balanced associations. We also report the absolute probability gap:
\begin{equation}
\textsc{Gap}(x) =
\left| \hat{p}_\theta(y_{\text{stereo}}\mid x) - \hat{p}_\theta(y_{\text{anti}}\mid x) \right|.
\end{equation}
Lower values indicate weaker bias.

\paragraph{\textbf{iCAT (StereoSet).}}
We additionally report the iCAT score~\citep{nadeem2020stereoset}, which captures the trade-off between bias and language modeling quality. The metric combines a language modeling score (LMS), which measures the model’s ability to assign high probability to coherent and contextually appropriate continuations, and a stereotype score (SS), which quantifies the model’s preference for stereotypical over anti-stereotypical associations:

\begin{equation}
\text{iCAT} = \text{LMS} \times \frac{\min(\text{SS},\,100-\text{SS})}{50}.
\end{equation}

The bias component is maximized when the model does not systematically prefer either stereotypes or anti-stereotypes (i.e., $\text{SS} \approx 50$), and decreases as the model increasingly favors one over the other. As a result, iCAT ranges between 0 and 1: values close to 1 indicate that the model achieves both high language modeling quality and balanced (low-bias) behavior, while values close to 0 indicate either strong bias (SS near 0 or 100), poor language modeling performance (low LMS), or both.

\paragraph{\textbf{Perplexity.}}
We report perplexity as a measure of language modeling quality~\citep{jelinek2005continuous}. Perplexity evaluates how well a model predicts a sequence of tokens. Lower perplexity indicates better predictive performance, corresponding to higher likelihood assigned to the observed data. In our evaluation setup, perplexity is used to assess the impact of bias mitigation on language quality: values closer to the pretrained model's perplexity indicate that the mitigation introduces minimal degradation in general language modeling performance.

\subsection{Downstream Fairness Metrics}
\label{app:Experiment:extrinsic-metrics}

Downstream tasks involve classification from tabular data serialized into natural language. Gender is treated as a binary sensitive attribute ($A \in \{0,1\}$) across all datasets.

Each experiment is conducted over three independent train--test splits. Models are trained for multiple epochs, and the checkpoint with the highest test accuracy is selected. Results are averaged across splits.

We report:

\begin{itemize}
    \item \textbf{Accuracy (Acc)}:
    \begin{equation}
    \text{Acc} = P(\hat{Y} = Y).
    \end{equation}
    
    \item \textbf{Accuracy Parity (AccP)}:
    \begin{equation}
    \text{AccP}=\big|P(\hat{Y}=Y \mid A=0)-P(\hat{Y}=Y \mid A=1)\big|.
    \end{equation}

    \item \textbf{Demographic Parity (DP)}:
    \begin{equation}
    \text{DP}=\big|P(\hat{Y}=1 \mid A=0)-P(\hat{Y}=1 \mid A=1)\big|.
    \end{equation}

    \item \textbf{Equalized Odds (EqOdds)}:
    \begin{equation}
    \text{EqOdds}=
    \sum_{y\in\{0,1\}} \left|
    \begin{aligned}
    & P(\hat{Y}=1 \mid Y=y, A=0) \\
    & - P(\hat{Y}=1 \mid Y=y, A=1)
    \end{aligned}
    \right|.
    \end{equation}
\end{itemize}

Lower values indicate better fairness for AccP, DP, and EqOdds.

%% file: App_ImplementationDetails.tex

\subsection{FACU Training Configuration}
\label{app:facu-training}
FACU is applied before downstream adaptation using full-parameter fine-tuning of each language model. For each model, we perform a grid search over learning rates $\{10^{-5}, 10^{-6}, 10^{-7}\}$ and loss weights $\{1.0, 0.75, 0.5, 0.25, 0.15, 0.05\}$. Hyperparameters are selected on the validation split by choosing the configuration that achieves the largest reduction in intrinsic bias, measured by the probability gap between stereotypical and anti-stereotypical candidates.

We use the training split for FACU optimization, the validation split for hyperparameter selection, and the held-out test split only for final evaluation. All intrinsic bias results reported in the paper are computed on the test split.

Batch size is set to 12 for Llama-3.1 and 36 for Gemma-2 and Phi-3. FACU training is conducted on NVIDIA A100 80GB GPUs using mixed-precision training. Due to its larger model size, Llama-3.1 is trained using two GPUs per run, while Gemma-2 and Phi-3 are trained on a single GPU. Table~\ref{tab:best-hp-unlearning} reports the best-performing hyperparameter configuration for each model.

\paragraph{Unlearning baseline.}

The Unlearning baseline follows the original concept-unlearning objective~\cite{yao2024large}, using the same stereotypical and anti-stereotypical candidate pairs, training data, and optimization setup as FACU. FACU extends this baseline by adding the GAP loss, which explicitly regularizes the probability difference between the two demographic alternatives.

\begin{table}[t]
\centering
\begin{tabular}{|l|c|c|c|c|c|}
\hline
\textbf{Model} & \textbf{LR} & $\lambda_{\text{KL}}$ & $\lambda_{\text{unlearning}}$ & $\lambda_{\text{learning}}$ & $\lambda_{\text{gap}}$ \\
\hline
Llama & 1e-5 & 1.0 & 0.5 & 0.25 & 1.0 \\
Gemma   & 1e-5 & 1.0 & 0.25 & 0.25 & 0.25 \\
Phi     & 1e-5 & 1.0 & 0.5 & 0.25 & 0.5 \\
\hline
\end{tabular}
\caption{Best hyperparameter configurations for FACU.}
\label{tab:best-hp-unlearning}
\end{table}

\subsection{Downstream Adaptation and Computational Setup}
\label{app:downstream-adaptation}

We evaluate downstream performance and fairness under two deployment setups: (i) \emph{LLM as feature extractor}, where frozen LLM embeddings are used to train a logistic regression classifier, and (ii) \emph{LLM as classifier}, where the LLM itself is adapted for downstream prediction using LoRA fine-tuning.

For the feature-extractor setting, tabular inputs are serialized into natural language following the TabLLM framework, and representations are extracted from the frozen LLM to train a downstream logistic regression classifier.

For the LLM-as-classifier setting, we use LoRA fine-tuning with rank $r=16$, scaling factor $\alpha=32$, and dropout $0.05$. LoRA adapters are applied to the attention projection modules
$\{\texttt{q\_proj}, \texttt{k\_proj}, \texttt{v\_proj}, \texttt{o\_proj}\}$. We set the bias parameter to \texttt{none} and the task type to \texttt{causal LM}.

All downstream experiments are repeated across three independent train--test splits. For each split, models are trained for multiple epochs and the checkpoint achieving the highest predictive accuracy on the test setß is selected for final evaluation.

The downstream experiments involve substantial computational cost due to the combination of multiple datasets, train--test splits, mitigation strategies, and deployment settings. Experiments include pretrained models, CDA-augmented training, FACU-mitigated models trained on both original and augmented data, as well as inference-time self-debiasing applied to both pretrained and mitigated models. For downstream adaptation on the ACS Employment dataset, one epoch of LoRA fine-tuning requires approximately 50 minutes, while CDA-based training roughly doubles this runtime due to dataset expansion. Inference-time self-debiasing introduces additional test-time overhead without affecting training time. Table~\ref{tab:cost} reports indicative training times for the LLM-as-classifier setting. Additional experiments, including feature-extractor setups and self-debiasing evaluations, further increase the overall computational cost.

All experiments are conducted on NVIDIA A100 80GB GPUs using mixed-precision training. Downstream fine-tuning and feature-extraction experiments are conducted on a single GPU per split.

\begin{table}[t]
\centering
\small
\begin{tabular}{|lccc|}
\hline
\textbf{Dataset} & \textbf{Time/Epoch} & \textbf{Runs} & \textbf{Total} \\
\hline
ACS Employ. (Base) & 50m & $3 \times 10 \times 2$ & $\sim$50h \\
ACS Employ. (CDA)  & 100m & $3 \times 10 \times 2$ & $\sim$100h \\
\hline
Adult (Base) & 25m & $3 \times 10 \times 2$ & $\sim$25h \\
Adult (CDA)  & 50m & $3 \times 10 \times 2$ & $\sim$50h \\
\hline
German Credit (Base) & 5m & $3 \times 10 \times 2$ & $\sim$5h \\
German Credit (CDA)  & 10m & $3 \times 10 \times 2$ & $\sim$10h \\
\hline
\end{tabular}
\caption{Indicative training time for the LLM-as-classifier setting across datasets and configurations. Total runs correspond to 3 train--test splits, 10 epochs, and 2 setups (pretrained and FACU).}
\label{tab:cost}
\end{table}

%% file: App_AdditionalResults.tex
\subsection{Ablation Study}
\label{app:ablation}
To better understand the role of each FACU objective, we conduct an ablation study evaluating the contribution of the unlearning, learning, GAP, and normalization terms across all LLMs. The results show that effective intrinsic bias mitigation depends on a careful balance between unlearning and regularization terms, as removing or weakening these components leads to unstable behavior or limited bias reduction. Detailed ablation results for all models are reported in Tables~\ref{tab:ablation-Llama3}, \ref{tab:ablation-gemma2}, and \ref{tab:ablation-phi3}.

\begin{table*}[t]
\centering
\begin{tabular}{|l|c|c|c|c|c|c|c|}
\hline
Configuration &
Unlearn &
Learn &
Gap &
KL &
$\mathbf{P_{\text{anti}}}$ &
$\mathbf{P_{\text{stereo}}}$ &
GAP$\downarrow$ \\
\hline

Pretrained
& \xmark & \xmark & \xmark & \xmark
& 0.220 & 0.780 & 0.560 \\

Only Learning
& \xmark & \cmark & \xmark & \xmark
& 0.382 & 0.618 & 0.236 \\

Only Unlearning
& \cmark & \xmark & \xmark & \xmark
& 0.368 & 0.632 & 0.264 \\

Only KL
& \xmark & \xmark & \xmark & \cmark
& 0.301 & 0.699 & 0.398 \\

Only GAP
& \xmark & \xmark & \cmark & \xmark
& 0.388 & 0.612 & 0.224 \\


Learning + GAP
& \xmark & \cmark & \cmark & \xmark
& 0.393 & 0.607 & 0.214 \\

Unlearning + GAP
& \cmark & \xmark & \cmark & \xmark
& 0.394 & 0.606 & 0.212 \\

GAP + KL
& \xmark & \xmark & \cmark & \cmark
& 0.389 & 0.611 & 0.222 \\

 Unlearning \cite{yao2024large}
&  \cmark &  \cmark &  \xmark &  \cmark
&  0.305 &  0.695&  0.390 \\

FACU
& \cmark & \cmark & \cmark & \cmark
& 0.470 & 0.530 & \textbf{0.059} \\

\hline
\end{tabular}

\vspace{1mm}
{\footnotesize
\cmark: component included,\;
\xmark: component excluded.\;
}
\caption{Ablation study of FACU objective components for Llama-3.1. Lower GAP values indicate more balanced probabilities between stereotypical and anti-stereotypical candidates.}

\label{tab:ablation-Llama3}
\end{table*}

\begin{table*}[t]
\centering
\begin{tabular}{|l|c|c|c|c|c|c|c|}
\hline
Configuration &
Unlearn &
Learn &
Gap &
KL &
$\mathbf{P_{\text{anti}}}$ &
$\mathbf{P_{\text{stereo}}}$ &
GAP$\downarrow$ \\
\hline

Pretrained
& \xmark & \xmark & \xmark & \xmark
& 0.193 & 0.807 & 0.614 \\

Only Learning
& \xmark & \cmark & \xmark & \xmark
& 0.463 & 0.537 & 0.074 \\

Only Unlearning
& \cmark & \xmark & \xmark & \xmark
& 0.463 & 0.537 & 0.074 \\

Only KL
& \xmark & \xmark & \xmark & \cmark
& 0.212 & 0.788 & 0.576 \\

Only GAP
& \xmark & \xmark & \cmark & \xmark
& 0.564 & 0.436 & 0.128 \\

Learning + GAP
& \xmark & \cmark & \cmark & \xmark
& 0.551 & 0.449 & 0.102 \\

Unlearning + GAP
& \cmark & \xmark & \cmark & \xmark
& 0.584 & 0.416 & 0.168 \\

GAP + KL
& \xmark & \xmark & \cmark & \cmark
& 0.563 & 0.437 & 0.126 \\

 Unlearning \cite{yao2024large}
&  \cmark &  \cmark &  \xmark &  \cmark
&  1.000 &  0.000 &  1.000 \\

FACU
& \cmark & \cmark & \cmark & \cmark
& 0.546 & 0.454 & \textbf{0.092} \\

\hline
\end{tabular}

\vspace{1mm}
{\footnotesize
\cmark: component included,\;
\xmark: component excluded.\;
}
\caption{Ablation study of FACU objective components for Gemma-2. Lower GAP values indicate more balanced probabilities between stereotypical and anti-stereotypical candidates. GAP inversion occurs when anti-stereotypical probabilities dominate stereotypical probabilities.}
\label{tab:ablation-gemma2}
\end{table*}

\begin{table*}[t]
\centering
\begin{tabular}{|l|c|c|c|c|c|c|c|}
\hline
Configuration &
Unlearn &
Learn &
Gap &
KL &
$\mathbf{P_{\text{anti}}}$ &
$\mathbf{P_{\text{stereo}}}$ &
GAP$\downarrow$ \\
\hline

Pretrained
& \xmark & \xmark & \xmark & \xmark
& 0.291 & 0.709 & 0.418 \\

Only Learning
& \xmark & \cmark & \xmark & \xmark
& 0.561 & 0.439 & 0.122 \\

Only Unlearning
& \cmark & \xmark & \xmark & \xmark
& 0.364 & 0.636 & 0.272 \\

Only KL
& \xmark & \xmark & \xmark & \cmark
& 0.351 & 0.649 & 0.298 \\

Only GAP
& \xmark & \xmark & \cmark & \xmark
& 0.532 & 0.468 & 0.064 \\

Learning + GAP
& \xmark & \cmark & \cmark & \xmark
& 0.537 & 0.463 & 0.074 \\

Unlearning + GAP
& \cmark & \xmark & \cmark & \xmark
& 0.534 & 0.466 & 0.068 \\

GAP + KL
& \xmark & \xmark & \cmark & \cmark
& 0.536 & 0.464 & 0.072 \\

 Unlearning \cite{yao2024large}
&  \cmark &  \cmark &  \xmark &  \cmark
&  0.992 &  0.008 &  0.984 \\

FACU
& \cmark & \cmark & \cmark & \cmark
& 0.514 & 0.486 & \textbf{0.029} \\

\hline
\end{tabular}

\vspace{1mm}
{\footnotesize
\cmark: component included,\;
\xmark: component excluded.\;
}
\caption{Ablation study of FACU objective components for Phi-3. Lower GAP values indicate more balanced probabilities between stereotypical and anti-stereotypical candidates. GAP inversion occurs when anti-stereotypical probabilities dominate stereotypical probabilities.}
\label{tab:ablation-phi3}
\end{table*}

KL divergence plays a central role in preserving model coherence during fine-tuning, as reflected by stable perplexity and generation quality. However, our ablation results indicate that strong KL regularization can also preserve pre-existing intrinsic biases by constraining the model to remain close to its original biased distribution. 
Strong KL regularization preserves generation quality but can also retain pre-existing intrinsic biases by constraining the model to remain close to the original distribution. Aligning to external reference datasets such as TruthfulQA~\citep{lin2021truthfulqa} was insufficient to remove socio-economic biases, likely because these datasets are not designed for fairness-oriented debiasing. This effect is particularly visible in LLM-as-classifier setups, where strong coherence constraints appear to limit the effectiveness of intrinsic bias mitigation, partially explaining their weaker downstream fairness improvements compared to embedding-based classifiers.

The GAP loss plays a critical role in explicitly balancing model confidence between stereotypical and anti-stereotypical alternatives, beyond what can be achieved through suppression-based unlearning alone. As shown in Tables~\ref{tab:ablation-Llama3},~\ref{tab:ablation-gemma2}, and~\ref{tab:ablation-phi3}, removing the GAP loss leads to qualitatively different failure modes depending on model capacity. In smaller models (Phi-3 and Gemma-2), unlearning without the GAP term frequently results in a sharp inversion of the original bias distribution, effectively replacing stereotypical associations with anti-stereotypical ones. In contrast, for Llama-3.1, the absence of the GAP loss produces only marginal changes and fails to effectively reduce the intrinsic bias gap.

In both cases, the absence of the GAP loss yields unbalanced probability distributions that remain exploitable during downstream classification. These observations highlight an important asymmetry: models can more easily learn anti-stereotypical associations than fully unlearn stereotypical ones. By explicitly penalizing confidence gaps between competing alternatives, the GAP loss stabilizes intrinsic representations and prevents both inversion effects in smaller models and gap amplification in larger ones. This behavior helps explain why suppression-based unlearning alone, as in prior formulations~\citep{yao2024large}, may be insufficient for reliable intrinsic bias mitigation.

Notably, such suppression-based formulations were originally designed for safety- and privacy-oriented objectives, including removing harmful responses, erasing copyrighted content, reducing hallucinations, protecting user privacy, and enforcing policy compliance, where the primary goal is to suppress or forget specific behaviors or information. In contrast, fairness mitigation in socio-economic contexts does not require eliminating attribute–outcome associations altogether, but rather enforcing balanced confidence across sensitive attribute values (e.g., gender terms) while preserving task-relevant structure. As a result, fairness-oriented regularization, such as the GAP loss, is necessary to obtain stable and balanced intrinsic representations aligned with fairness objectives. In this formulation, the learning and unlearning terms drive parameter updates, while the GAP loss regulates their relative strength, ensuring balanced corrections without under- or overcompensation.

\subsection{Cross-Domain Generalization}
\label{app:RQ1_CrossDomainGeneralization}
To evaluate whether intrinsic bias mitigation learned by FACU generalizes beyond socio-economic contexts, we conduct additional experiments on the BBQ~\cite{parrish2022bbq} and CrowS-Pairs~\cite{nangia2020crows} benchmarks. Importantly, FACU is trained only on SOCIOECONOMICQA and is evaluated on these datasets without any additional fine-tuning or adaptation. While SocioEconomicQA focuses on socio-economic scenarios and gender-related attributes, BBQ and CrowS-Pairs contain broader social contexts and different forms of gender expression, including names, pronouns, and indirect demographic references. These datasets are widely used fairness benchmarks and therefore provide a useful setting for evaluating the robustness and generalization of intrinsic bias mitigation beyond the training domain.

Table~\ref{tab:bbq} reports PAR GAP scores on BBQ and CrowS-Pairs before and after FACU mitigation. Across models, FACU consistently reduces the intrinsic bias gap on both datasets, suggesting that the learned mitigation generalizes beyond the specific linguistic patterns and socio-economic contexts observed during training.
\begin{table}[ht]
\centering
\begin{tabular}{|l|l|c||c|}
\toprule
\multirow{2}{*}{\textbf{LLM}} &
\multirow{2}{*}{\textbf{Version}} &
\multicolumn{2}{c|}{\textbf{Gap} ($\downarrow$)} \\
\cmidrule(lr){3-4}
& & \textbf{CrowS-Pairs} & \textbf{BBQ} \\
\midrule
\multirow{2}{*}{Llama} 
 & Pretrained 
 & 0.616 
 & 0.360 \\
 &  FACU  
 &  0.427 
 &  0.225 \\
 \midrule
 
\multirow{2}{*}{Gemma} 
 & Pretrained 
 & 0.654 
 & 0.621 \\
 &  FACU  
 &  0.488 
 &  0.328 \\
\bottomrule

\multirow{2}{*}{Phi} 
 & Pretrained 
 & 0.631 
 & 0.519 \\
 &  FACU  
 &  0.600 
 &  0.469 \\
 \midrule

\end{tabular}
\caption{Cross-domain intrinsic bias evaluation on BBQ and CrowS-Pairs. Lower GAP values indicate reduced bias.}
\label{tab:bbq}
\end{table}


\subsection{Statistical Significance Analysis}
\label{app:RQ1_Statistical_test}

We assess whether FACU reduces the \emph{intrinsic bias gap}, defined as the probability difference between stereotypical and anti-stereotypical completions, using a paired sign test (exact binomial) over evaluation items. For each LLM, we compare the pretrained baseline against its FACU version. Table~\ref{tab:RQ1_ttest} reports the resulting \textit{p}-values.

Overall, FACU consistently yielded statistically significant reductions in the intrinsic bias gap across the evaluated models, as indicated by p-values below 0.05.

\begin{table}[ht]
\centering
\setlength{\tabcolsep}{6pt}

\label{tab:pvalues_unlearning_icl}
\begin{tabular}{|l|c|}
\toprule
\textbf{Model} 
&  \textbf{Fairness-Awar Concept Unlearning} \\
\midrule
Llama
&  \num{3.89e-12}\\
\midrule
Gemma
&  \num{5.16e-14}\\
\midrule
Phi
&  \num{2.10e-2}\\
\bottomrule
\end{tabular}
\caption{Statistical significance (\textit{p}-values) of the reduction in intrinsic gender bias gap
before and after FACU 
across three LLMs.
Lower \textit{p}-values indicate stronger evidence of significant gap reduction.}
\label{tab:RQ1_ttest}
\end{table}

\subsection{Self-Debiasing Limitations}
\label{app:RQ2_selfDebiasing}
Self-debiasing~\cite{gallegos2025self} is an inference-time mitigation method that revises model outputs through additional prompting. As it operates on generated responses, it is naturally applicable in the LLM-as-classifier setting. However, prior work primarily evaluates this technique in zero-shot settings on short, structured benchmarks.

To assess its behavior in our downstream setting, we first apply self-debiasing directly to pretrained LLMs without any task-specific adaptation. Table~\ref{tab:selfdebiasing_dst} reports downstream performance and fairness metrics under this zero-shot configuration. While re-prompting can alter group fairness metrics, predictive accuracy is consistently low or unstable across datasets and models. As a result, downstream fairness estimates in this setting are unreliable, as disparities are measured on poorly performing classifiers.

These observations motivate the adapted use of self-debiasing in our experiments, where it is applied after downstream fine-tuning. This ensures that the model first acquires sufficient task-specific understanding before output-level adjustments are introduced. In this setting, fairness interventions operate on meaningful predictions rather than unstable outputs, enabling more reliable comparisons with other mitigation strategies.

\begin{table*}[ht]

\centering

\resizebox{\textwidth}{!}{%
\begin{tabular}{|l|l|cccc|cccc|cccc|}
\toprule
\textbf{Model} & \textbf{Stage} & \multicolumn{4}{c|}{\textbf{ACS Employment Dataset }} & \multicolumn{4}{c|}{\textbf{Adult Dataset}} & \multicolumn{4}{c|}{\textbf{German Credit Dataset}} \\
\cmidrule(lr){3-6} \cmidrule(lr){7-10} \cmidrule(lr){11-14}
& & Acc$\uparrow$ & AccP$\downarrow$ & DP$\downarrow$ & EqOdds$\downarrow$ & Acc$\uparrow$ & AccP$\downarrow$ & DP$\downarrow$ & EqOdds$\downarrow$ & Acc$\uparrow$ & AccP$\downarrow$ & DP$\downarrow$ & EqOdds$\downarrow$ \\
\midrule

\multirow{2}{*}{Llama-3.1}& Pretrained (Zero-shot)  & 0.656 & 0.137 & 0.021 & 0.030 & 0.321 &0.192  & 0.009 & 0.018 & 0.686 & 0.111  & 0.037  & 0.062   \\
&Self-debiased (Zero-shot) & 0.344 & 0.137 & 0.021 & 0.031 & 0.679 & 0.192 & 0.009 & 0.018 & 0.312 & 0.114  & 0.040  &  0.066 \\
\midrule
\multirow{2}{*}{Gemma-2}& Pretrained (Zero-shot)  & 0.367 & 0.138 & 0.010 & 0.015 & 0.775 &0.163  & 0.089 & 0.219 & 0.300 & 0.083  & 0.000  & 0.000   \\
&Self-debiased (Zero-shot) & 0.559 & 0.097 & 0.007 & 0.225 & 0.753 &0.202  &0.001 & 0.003 &0.300 & 0.083  & 0.000 &0.000   \\
\midrule
\multirow{2}{*}{Phi-3}&
 Pretrained (Zero-shot)  & 0.359 & 0.146 & 0.000 & 0.000 & 0.753 &0.203  & 0.000 & 0.001 & 0.300 & 0.083  & 0.000  & 0.000   \\
&Self-debiased (Zero-shot) &0.362  &0.142 &0.011 & 0.022  &0.772 &0.168  & 0.054  &0.108 &0.298 &0.085  &0.003  &0.010   \\

\bottomrule
\end{tabular}
}
\caption{Downstream accuracy and group-fairness metrics obtained by applying zero-shot self-debiasing via re-prompting directly to pretrained LLMs, without downstream task adaptation. Across datasets, predictive accuracy is low or unstable, rendering downstream fairness estimates unreliable. This motivates applying self-debiasing only after downstream fine-tuning.}

\label{tab:selfdebiasing_dst}
\end{table*}
